\documentclass[letterpaper, 10 pt, conference]{ieeeconf}  

\IEEEoverridecommandlockouts                              
\usepackage[utf8]{inputenc}
\usepackage[T1]{fontenc}

\usepackage{amsmath}
\usepackage{amssymb}
\usepackage{graphicx}
\usepackage[hidelinks]{hyperref}
\usepackage{float}
\usepackage{bbm}

\newcommand{\bs}{\boldsymbol}

\newcommand{\bth}{\bs{\theta}}

\newcommand{\Figure}[1]{Figure~\ref{#1}}
\newcommand{\Table}[1]{Table~\ref{#1}}
\newcommand{\Section}[1]{Section~\ref{#1}}

\title{\LARGE \bf
Fully Convolutional Networks for Dense Water Flow Intensity Prediction in Swedish Catchment Areas
}

\author{Aleksis Pirinen$^1$, Olof Mogren$^1$ and Mårten Västerdal$^2$ \\
$^1$RISE Research Institutes of Sweden\\
$^2$Department of City Planning and Sustainability, Kungsbacka municipality\\
\tt\small \{aleksis.pirinen@ri.se, olof.mogren@ri.se, marten.vasterdal@kungsbacka.se\}}

\begin{document}

\maketitle
\thispagestyle{empty}
\pagestyle{empty}

\begin{abstract}
Intensifying climate change will lead to more extreme weather events, including heavy rainfall and drought. Accurate stream flow prediction models which are adaptable and robust to new circumstances in a changing climate will be an important source of information for decisions on climate adaptation efforts, especially regarding mitigation of the risks of and damages associated with flooding. In this work we propose a machine learning-based approach for predicting water flow intensities in inland watercourses based on the physical characteristics of the catchment areas, obtained from geospatial data (including elevation and soil maps, as well as satellite imagery), in addition to temporal information about past rainfall quantities and temperature variations. We target the one-day-ahead regime, where a fully convolutional neural network model receives spatio-temporal inputs and predicts the water flow intensity in every coordinate of the spatial input for the subsequent day. To the best of our knowledge, we are the first to tackle the task of \emph{dense} water flow intensity prediction; earlier works have considered predicting flow intensities at a sparse set of locations at a time. An extensive set of model evaluations and ablations are performed, which empirically justify our various design choices. Code and preprocessed data have been made publicly available at \url{https://github.com/aleksispi/fcn-water-flow}.
\end{abstract}

\section{INTRODUCTION}
As climate change intensifies, hydrological conditions will change. This will manifest itself both in the form of water shortages and as flooding in cases of intense precipitation. According to the Swedish Environmental Protection Agency, the climate in Sweden is becoming warmer and wetter \cite{bernes2017varmare}, and municipalities are encouraged to increase their climate adaptation efforts, especially regarding mitigating the risks of, and damages associated with, flooding \cite{mossberg2015varmens,schultze2022national}. The effects of climate change on rainfall-runoff will be more severe further north \cite{salmonsson2014assessing}. At the same time, the hydrological conditions in Sweden have been severely disturbed during the last two hundred years, with wetlands being drained and natural streams being straightened, which will further increase the effects of extreme weather events.

Hydrological modeling can shed light on the dynamics of water flow and how it is affected by various aspects of the environment. This can in turn allow for making informed decisions about the efficacy of nature-based climate change adaptation techniques such as wetland restoration, urban greening, and soil protection. Traditional hydrological models are based on expert knowledge and physical properties such as the preservation of volume, which have to be specified a priori. These work well for a certain domain if they are properly calibrated, but have difficulties generalizing to wider environmental categories \cite{tsai2021calibration}. Statistical data driven modeling, including machine learning (ML), is an alternative which has the potential to become more robust if it can be trained on a large enough dataset with a suitable learning signal. This way, not only can the flow intensity be estimated for any given water course following a heavy precipitation event, but also the response time of the given area, i.e.~an estimation of the time lapse from the precipitation event to peak flow. Such information is vital to better understand flood risks and the effects of flood and drought mitigation, as well as general hydrological implications from changes in land use.

In this work we propose an ML-based approach for water flow intensity prediction that leverages the physical characteristics of a catchment area. We target the one-day-ahead regime, where a fully convolutional neural network \cite{long2015fully} receives spatio-temporal inputs and predicts the water flow intensity at every coordinate for the subsequent day (the same modeling should however be able to handle other time horizons with minor modifications). Two important novelties of our proposed approach are:
\begin{itemize}
    \item In addition to temporal data (past rainfall and temperatures), we include spatial data as inputs to the modeling, provided as satellite imagery and several derived GIS layers. This allows the model to build internal representations about relationships between temporal and spatial aspects of the local environment (including land cover, soil depth and moisture, and elevation).
    \item Using a fully convolutional model, we tackle the task of \emph{dense} water flow intensity prediction, as opposed to only predicting flow intensities for a sparse set of spatial locations. To the best of our knowledge, we are the first to consider the dense prediction task.
\end{itemize}

The remainder of this paper is organized as follows. In \Section{sec:rel-work} we provide a brief overview of the related work. Then, in \Section{sec:dataset}, we describe in detail the data we have used for modeling, training and evaluation. In \Section{sec:method} we explain our approach for tackling the water flow intensity prediction task, and our proposed approach is empirically evaluated against alternative methods in \Section{sec:results}. Finally, the paper is concluded in \Section{sec:conclusion}.

\section{RELATED WORK}\label{sec:rel-work}
Water flow prediction (also known as \textit{stream forecasting}, or \textit{rainfall-runoff modeling}) for rivers in the U.S. have been modeled using Long Short-Term Memory (LSTM \cite{hochreiter1997long}) networks \cite{jia2021physics,frame2021post,frame2022deep,gauch2021proper}. The modeling follows a traditional setup inspired by earlier physics-based hydrological models such as the U.S. National Water Model (NWM), based on WRF-Hydro \cite{cosgrove2015hydrologic}. Jia et al. \cite{jia2021physics} modeled river segments using an LSTM network with graph convolutions. One segment in the river network corresponds to a distance that the water flows during approximately one day. Input features include daily average precipitation, daily average air temperature, date of the year, solar radiation, shade fraction, potential evapotranspiration, elevation, length, slope, and width of each segment. Models were trained using a physics-informed setup where a traditional flow model acted as a teacher for the machine learning model. LSTM networks have also been used for post-processing the output from the NWM \cite{frame2021post}. Similar to our work, most of these prior works have focused on next-day predictions. However, there are examples of hourly predictions \cite{gauch2021proper}.

Others have also employed convolutional neural networks (CNNs) for stream forecasting \cite{duan2020using,van2020deep,oliveira2023streamflow}. However, in contrast to us, these works do not incorporate spatial data from satellites or GIS, but instead model only the much lower-dimensional data (in single coordinates or very small neighborhoods, not entire areas as in our setup) provided as a feature vector for each time step, similar to the models using LSTMs.
More broadly, deep learning has been used for many related tasks, such as groundwater level estimation \cite{wunsch2021groundwater}, water quality estimation \cite{varadharajan2022can}, and rainfall-runoff \cite{kratzert2018rainfall}.

While some of the above mentioned works on stream flow estimation include information about the near environment (such as elevation and slope), none of them use detailed spatial information inputs as is proposed in our work. The use of fully convolutional neural networks to encode this information, in combination with traditional inputs such as rainfall and temperature, has the potential of representing more complex relationships and can result in a more detailed view of the near environment. It also enables us to perform \emph{dense} water flow intensity prediction, different to prior works.

\section{DATASET DESCRIPTION}\label{sec:dataset}
\begin{figure}[h]
    \centering
    \frame{\includegraphics[width=0.75\columnwidth]{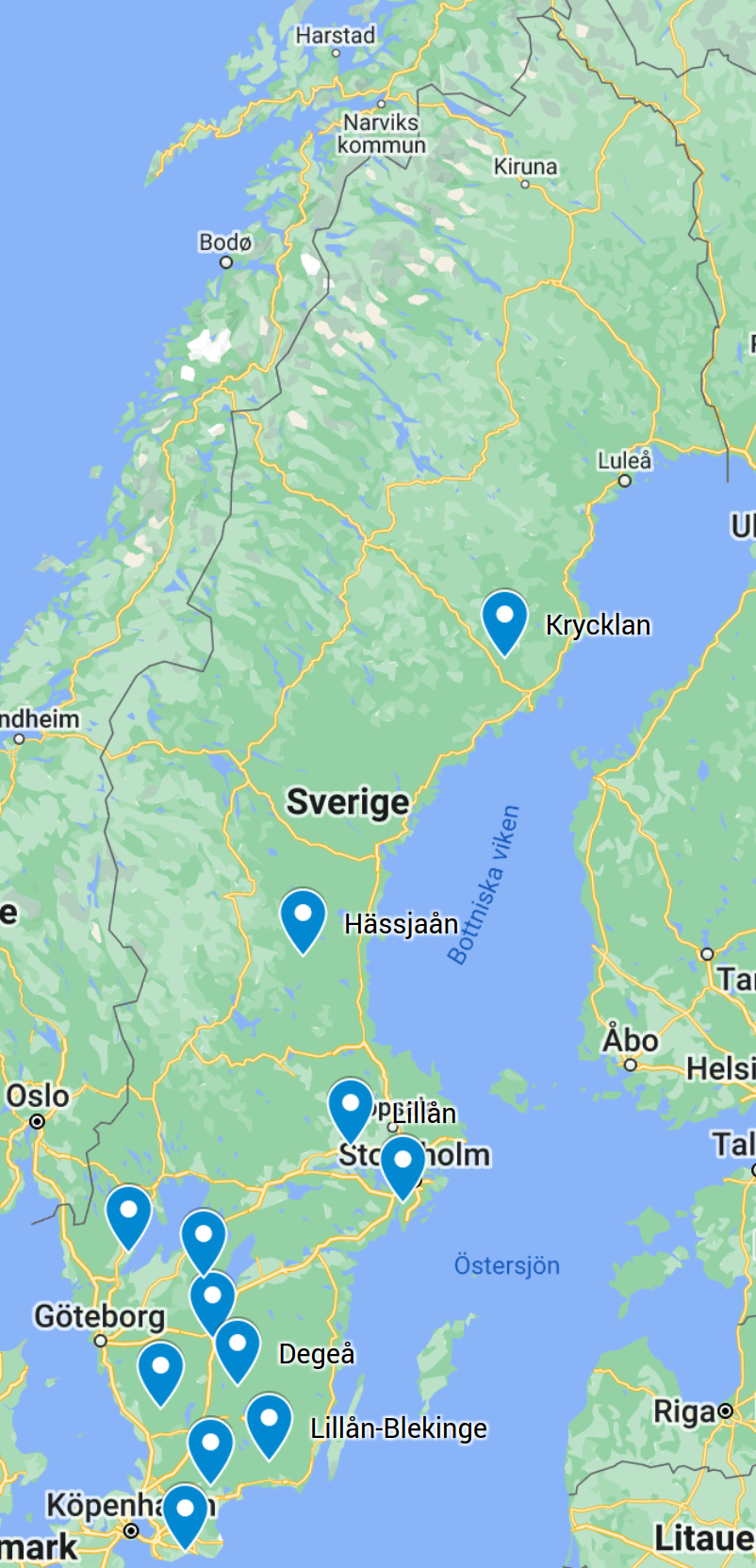}}
    \caption{The data used in this paper comes from 12 locations in Sweden: Jököping (Tabergsån), Knislinge (Almaån), Krycklan, Skivarp (Skivarpsån), Skövde (Ösan), Torup (Kilan), Tumba (Saxbroån), Dalbergsån, Degeå, Hässjaån, Lillån, and Lillån-Blekinge. Note that Lillån and Lillån-Blekinge are far apart (Lillån is north-west of Stockholm).}
    \label{fig:locations}
\end{figure}

We use data from 12 locations in Sweden, based on where the Swedish Meteorological and Hydrological Institute (SMHI) has stations for measuring weather and water flow data. These locations are Jönköping (Tabergsån), Knislinge (Almaån), Krycklan, Skivarp (Skivarpsån), Skövde (Ösan), Torup (Kilan), Tumba (Saxbroån), Dalbergsån, Degeå, Hässjaån, Lillån, and Lillån-Blekinge
(see \Figure{fig:locations}).

\begin{figure*}[ht!]
    \centering
    \includegraphics[width=0.99\textwidth]{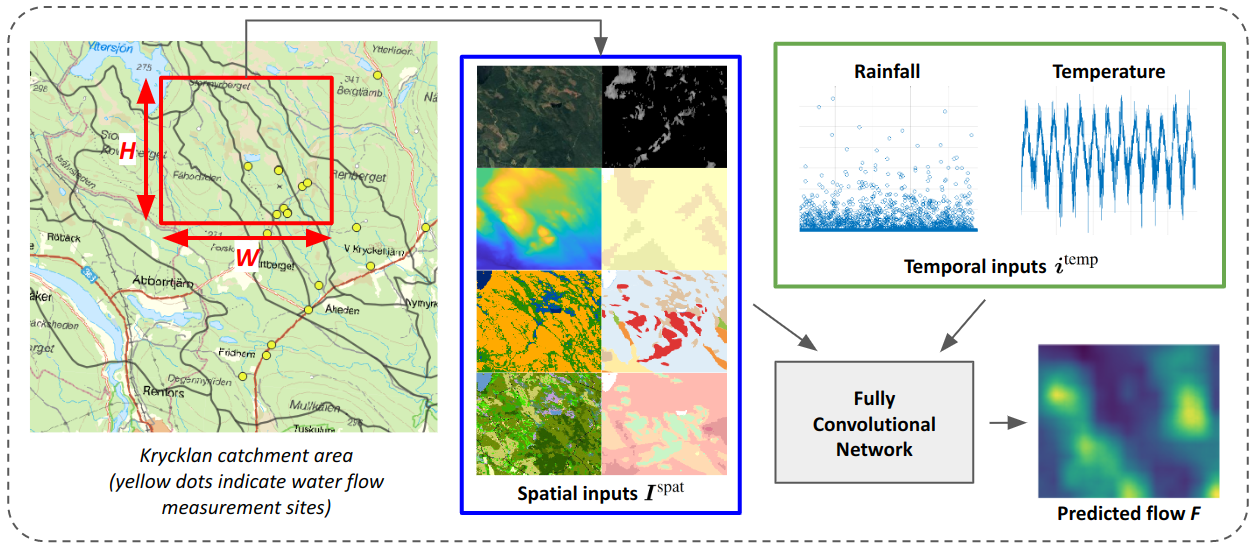}
    \caption{Overview of our machine learning (ML) approach for dense water flow intensity prediction in catchment areas. The fully convolutional neural network receives both spatial and temporal inputs and produces a map of predicted water flow intensities. The spatial input $\bs{I}^\text{spat} \in \mathbb{R}^{H \times W \times C}$ represents relevant properties of the region (red rectangle, which can be located anywhere) in which to perform water flow intensity prediction (e.g.~elevation map, soil moisture, land cover). The temporal inputs $\bs{i}^\text{temp} = \{r_j, \tau_j\}_{j=t-T}^{j=t-1}$ are the daily average rainfall amounts $r_j$ and temperatures $\tau_j$ over the past $T$ days, provided at times $t-T, t-T+1, \dots, t-1$. The model then produces the flow map $\bs{F} \in \mathbb{R}^{H \times W}$, where each pixel $(k, l)$ represents the model's estimate of the daily average water flow intensity at location $(k, l)$ in the input map during day $t$. Note that our model is only trained in a very sparse set of coordinates (since water flow intensity measurement stations are very scarcely located), but the model nonetheless predicts flow intensities in every coordinate, not only in those for which contain flow intensity measurement stations.}
    \label{fig:method-overview}
\end{figure*}

In each location we have access to the following spatial data layers (see also \Figure{fig:method-overview}):
\begin{itemize}
    \item satellite RGB image (Sentinel-2) from the Land Survey of Sweden (Lantmäteriet), $10\text{m} \times 10\text{m}$ resolution;
    \item elevation map from the Land Survey of Sweden, $50\text{m} \times 50\text{m}$ resolution;
    \item terrain slope map from the Land Survey of Sweden, $50\text{m} \times 50\text{m}$ resolution;
    \item soil moisture map the Swedish University of Agricultural Sciences, $2\text{m} \times 2\text{m}$ resolution;
    \item land cover map the Swedish Environmental Protection Agency, $10\text{m} \times 10\text{m}$ resolution;
    \item soil type map from the Geological Survey of Sweden, $10\text{m} \times 10\text{m}$ resolution;
    \item soil depth map from the Geological Survey of Sweden, $10\text{m} \times 10\text{m}$ resolution;
    \item hydraulic conductivity map from the Geological Survey of Sweden, $100\text{m} \times 100\text{m}$ resolution;
\end{itemize}
The elevation map provides each coordinate's elevation above the ocean level, whereas the terrain slope map provides the slope of each coordinate (obtained by computing differences between adjacent coordinates of the elevation map).

All input data were provided as spatially aligned sets of raster maps, each of size $825 \times 1244$ pixels.
For all locations except the Krycklan catchment area, there is exactly one water flow intensity measurement station. In Krycklan there are 14 such measurement stations, so in total there are 28 water flow intensity measurement stations in the dataset. In each measurement site, the daily average water flow intensity ($\text{m}^3 / \text{sec}$) is provided. The specific length of each time series varies, but generally span several decades (in average about three decades). 

In each location we also have access to two additional time series -- daily cumulative rainfall (mm) and daily average temperature (degrees Celsius). These were obtained from the Swedish Meteorological and Hydrological Insitute (SMHI), and are often measured some distance away from the water flow intensity measurement sites (typically 1-3 kilometers).

It is common with missing measurements in the time series. For those time series which are used as model inputs (see \Section{sec:method}), this is remedied by linearly interpolating the missing values between end points. Note that this is not done for the regressor (water flow intensity), since we only want the model to learn on and be evaluated on actual measurements.
\\ \\
\textbf{Data preprocessing.} We perform normalization of both the spatial and temporal data. Specifically, the spatial inputs are normalized to the $[0,1]$-range by dividing with the maximum value (layer-wise) across all locations. A similar $[0,1]$-normalization is performed for the temporal inputs (rainfall, temperature, and water flow intensity). We also tried another common normalization technique, where the variables are normalized to zero mean and unit variance, but empirically found that the $[0,1]$-normalization works best in our setup.

\section{METHODOLOGY}\label{sec:method}
In this section we provide an overview of the approach that we have developed for tackling the water flow intensity prediction task. See \Figure{fig:method-overview} for an overview of our model and setup.

Our model leverages both spatial inputs $\bs{I}^\text{spat} \in \mathbb{R}^{H \times W \times C}$ and temporal inputs $\bs{i}^\text{temp} = \{r_j, \tau_j\}_{j=t-T}^{j=t-1}$ (cf.~\Section{sec:dataset}), in order to predict water flow intensities $f_t$ at time\footnote{Similar to most prior works, we target next-day prediction, but the model and approach can be extended to predict further into the future.} $t$. The task of the model is to predict the water flow intensity $f_t^{h,w}$ at every coordinate $(h,w)$ in a given geographical area of size $H \times W$ given $\bs{I}^\text{spat}$ and $\bs{i}^\text{temp}$. For the temporal inputs, we have chosen to only use readily available rainfall $r_{t-T}, \dots, r_{t-1}$ and temperature data $\tau_{t-T}, \dots, \tau_{t-1}$ for the past $T$ days (with $T=20$ in our setup), and not water flow intensity data $f_{t-T}, \dots, f_{t-1}$, which is often unavailable in practice. In particular, note that water flow intensity is only measured at a very sparse subset of all coordinates in each location -- and there are many locations in Sweden (and beyond) where no such measurement setups exist at all. Hence, for the model to be useful in a much larger set of contexts, it does not rely on past water flow intensities as input. However, in \Section{sec:experiments} we also compare with model variants that include past water flow intensities when predicting future flow intensities.

The spatial input $\bs{I}^\text{spat} \in \mathbb{R}^{H \times W \times C}$ contains relevant information regarding land and topological properties that affect the water flow intensity in any given coordinate. These spatial input layers were introduced in \Section{sec:dataset}. In our setup we let $H=W=100$, which corresponds to a real-world area of size $1\text{km} \times 1\text{km}$. The number of layers $C$ is 10 in our case (three layers for the RGB satellite images, and one layer each for the other types of spatial input). 

Since the task is to predict the water flow intensity in every coordinate in a map of size $H \times W$, based (in part) on spatial inputs of size $H \times W \times C=100 \times 100 \times 10$, we have opted for a fully convolutional neural network\footnote{We use the \emph{FCN8} model from the open-source FCN library \cite{pytorch-fcn2017}.} (FCN) \cite{long2015fully}. This architecture expects a spatial input at one end, and gives a spatial output at the other end. To achieve this, we first concatenate the spatial and temporal data $\bs{I}^\text{spat}$ and $\bs{i}^\text{temp}$ into a unified input $\bs{I} \in \mathbb{R}^{H \times W \times (C+2T)}$. The first $C$ channels are identical to $\bs{I}^\text{spat}$, while the last $2T$ channels are obtained by tiling rainfall $r_{t-T}, \dots, r_{t-1}$ and temperature data $\tau_{t-T}, \dots, \tau_{t-1}$ into $H \times W$-dimensional maps that are concatenated along the channel-dimension (each such map contains $HW$ copies of a single value $r_i$ or $\tau_i$, for $i \in \{t-T, \dots, t-1\}$). Given an input $\bs{I}$, the water flow intensity mapping is straightforward: $\bs{F} = g_{\bth}(\bs{I})$, where $\bth$ denotes the learnable parameters of the FCN $g$. We also evaluate and compare with other architecture variants in \Section{sec:experiments}.

\subsection{Model Training}\label{sec:training}
We randomly set aside 9 of the 12 data locations for training and 3 for validating the models. Specifically, the models are evaluated in Jönköping (Tabergsån), Hässjaån and Lillån, and trained on the other 9 locations; please refer to \Section{sec:dataset} for details about the dataset. In particular, note that Lillån-Blekinge is in the training set, but it is at a vastly different location than the validation set location Lillån (cf.~\Figure{fig:locations}).

\begin{figure}[t]
    \centering
    \includegraphics[width=0.40\textwidth]{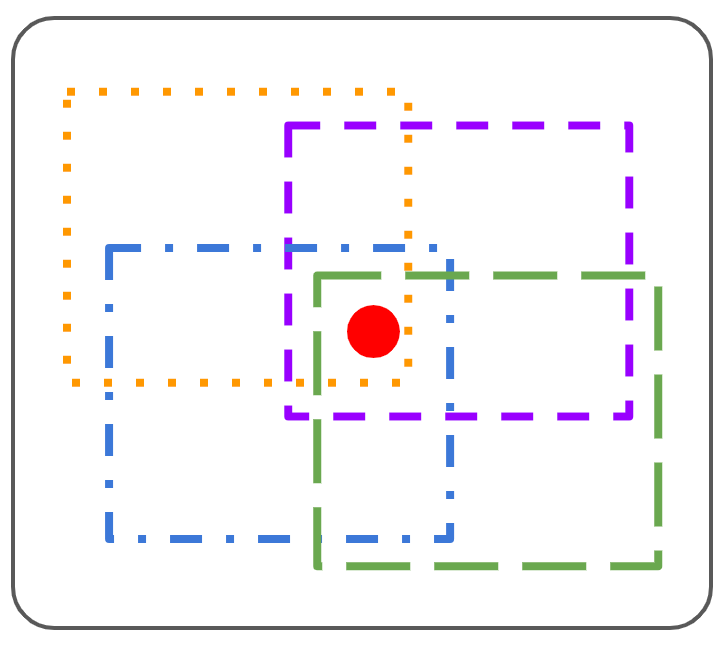}
    \caption{Examples of possible sampled spatial locations (colored, dashed rectangles) that contain a water flow intensity measurement station (red dot). The random sampling increases the data variability (compared to e.g.~always requiring the measurement station to be at the center). Since each possible spatial location has size $H \times W$, with $H=W=100$ in our setup, the union of all possible such rectangles covers roughly $200 \times 200$ pixels, which corresponds to a real-world area of size $2\text{km} \times 2\text{km}$.}
    \label{fig:positions}
\end{figure}

Each training input $\bs{I}$ in a batch is generated by randomly sampling a location and time period from the training set. Once a specific location has been randomly selected, we randomly sample a sub-region of size $H \times W$ which contains a water flow measurement site at the given location -- see \Figure{fig:positions}. This results in the spatial input $\bs{I}^\text{spat} \in \mathbb{R}^{H \times W \times C}$. After having sampled a spatial location, we then concatenate the temporal information $\bs{i}^\text{temp}$ from a randomly sampled time interval (of $T$ consecutive days), to obtain the input $\bs{I} \in \mathbb{R}^{H \times w \times (C + 2T)}$.

There are roughly $25 \cdot 100 \cdot 100=250,000$ different spatial training inputs (25 measurement sites, and at each site there are roughly $100 \cdot 100$ possible locations for an enclosing rectangle of size $H \times W = 100 \times 100$ -- see \Figure{fig:positions}). Note however that there is a spatial overlap between all different rectangles at a given site, which significantly reduces the data variability, compared to if all $250,000$ different rectangles would have come from different locations. The sites have on average roughly three decades of daily rainfall, temperature and water flow averages, which means there are $30 \cdot 365 \approx 11,000$ different temporal inputs per site. Hence, in total there are roughly $250,000 \cdot 11,000 = 2.75 \cdot 10^9$ spatio-temporal training inputs. Again, however, note that there is an overlap between a large majority of these training inputs, so the training set is effectively much smaller.

For model evaluation (see \Section{sec:experiments}), the end objective is to minimize the root-mean-square error (RMSE) of the predicted water flow intensities (thus note that the error is measured in $\text{m}^3 / \text{sec}$), assessed based on ground truth flow intensities. During training, in order to balance loss smoothness with robustness to outliers, we use the Huber loss: 
\begin{equation}\label{eq:loss}
    \mathcal{L}(f, f^\text{gt}) = 
    \begin{cases}
        \frac{1}{2}\left( f - f^\text{gt} \right)^2 &\text{ if } \|f - f^\text{gt}\| \leq \delta \\
        \delta \left(\|f - f^\text{gt}\| - \frac{1}{2} \delta \right) &\text{ else}
    \end{cases}
\end{equation}
where $f$ and $f^\text{gt}$ denote predicted and ground truth flow intensity, respectively. We set $\delta=1$ by default, as it is shown to result in the best performance (see \Section{sec:experiments}, where we also compare with other loss functions). The Huber loss can be seen as a combination of the commonly used MSE- and L1-losses, where the MSE-loss is applied when the error is smaller than the threshold $\delta$, and the L1-loss is applied otherwise.

\begin{table*}[t]
\centering
\caption{Experimental results on the validation set for our main model, its ablated variants, and baselines. We report the root-mean-square error (RMSE; lower is better). Column 1 represents our main FCN model, cf.~\Section{sec:method}. Columns 2-4 represent model variants which omit some of the spatial input layers. Columns 4-6 represent variants which omit some temporal inputs. Columns 7-8 represent baseline methods against which to compare the results in columns 1-5. Note that \emph{PREVIOUS FLOW} leverages past water flow intensity information that is unavailable to the other approaches.}\label{table:main-results}
\scalebox{1.14}{
\begin{tabular}{|c||c|c|c||c|c|c||c|c|}
\hline
\textbf{Main model} & \textbf{No-elev} & \textbf{Only-elev} & \textbf{No-soil} & \textbf{No-temp} & \textbf{No-rain} & \textbf{Half-time-hist} &\textbf{Mean-per-site} & \textbf{Previous flow} \\ \hline \hline
1.35 & 4.74 & 3.22 & 4.49 & 1.91 & 5.91 & 1.40 & 2.05 & 0.59 \\ \hline
\end{tabular}}
\end{table*}

Note that for each predicted flow map $\bs{F}$, the loss \eqref{eq:loss} is only given for an extremely sparse set of coordinates (most commonly in a single point). This is because in the ground truth flow map $\bs{F}^\text{gt}$, we only have access to water flow measurements in very few coordinates (since the measurement sites are so sparsely located in the data). Despite this extreme loss sparsity, we show in \Section{sec:experiments} that the model generalizes well to unseen data. This finding is is in line with earlier works that have shown that it is possible to train semantic segmentation models from extremely few annotated pixels \cite{shin2021all}.

For model parameter optimization, we resort to Adam \cite{kingma2014adam} with batch size $64$ and learning rate $2\cdot 10^{-4}$. The model is trained for $250,000$ batches, which takes about 48 hours on the GPU-equipped (Titan V100) work station that is used for experimentation. To improve model generalization towards unseen data, we resort to the customary deep learning training technique of augmenting the data by horizontal and vertical flips of the inputs (an independent probability of $50$\% per flip).

\section{EXPERIMENTS}\label{sec:experiments}
In this section we present the results of our empirical model evaluations on the validation set. We first describe the various baselines and model variants in \Section{sec:baselines}. Then, in \Section{sec:results}, the empirical results are presented.

\begin{figure*}[h]
    \centering
\includegraphics[width=0.875\textwidth]{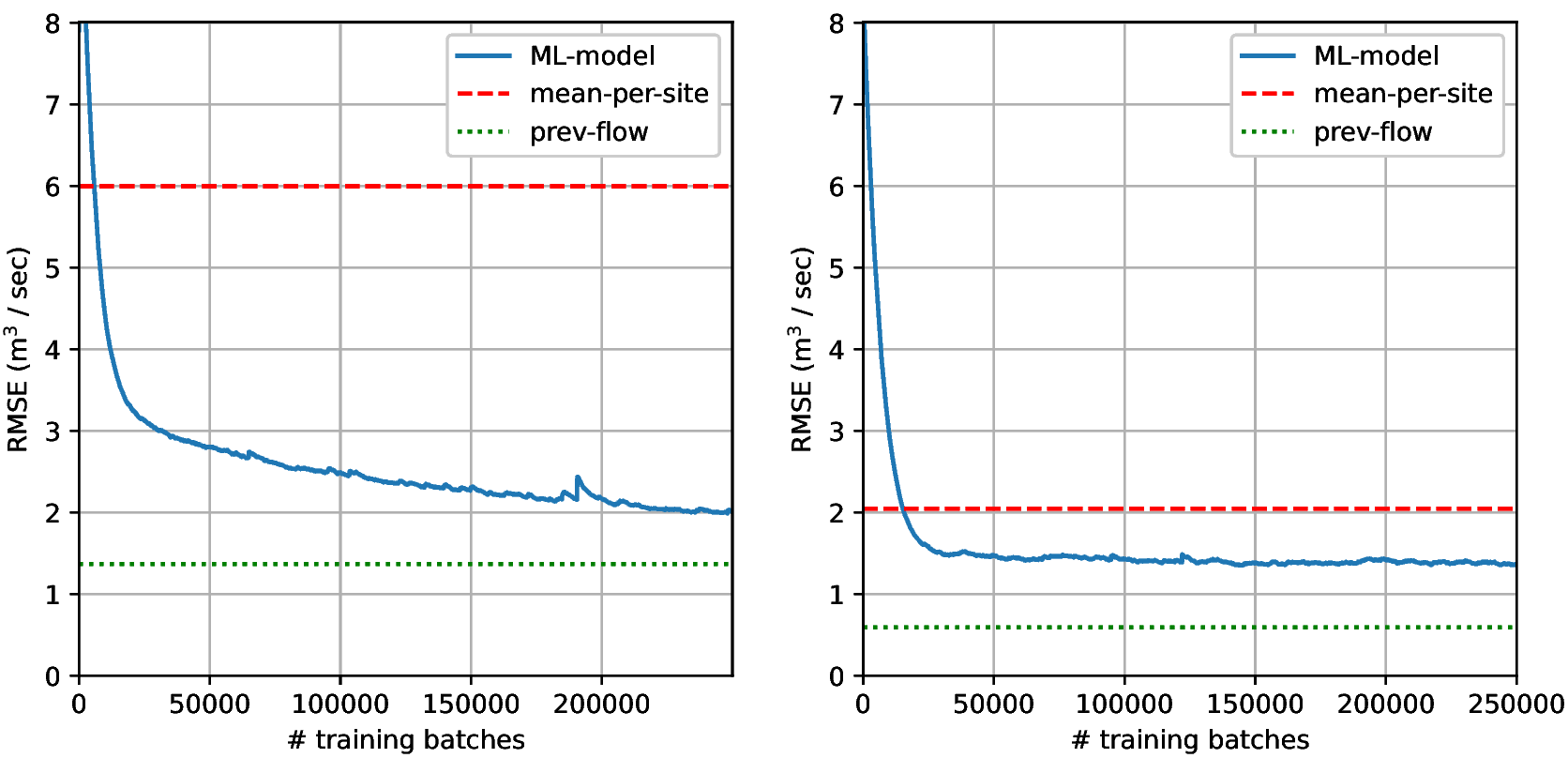}
    \caption{Training (left) and validation (right) RMSE curves during model training for our main ML model (FCN) described in \Section{sec:method}. It can be seen that the validation RMSE curve of our model flattens (marginally decreases) throughout training, i.e.~the model does not begin to overfit on the training data despite the relatively small size of the training set. \emph{Mean-per-site} and \emph{previous flow} represent baselines, of which \emph{previous flow} can be seen as an oracle that leverages past water flow intensity information that is unavailable to our model.}
    \label{fig:main-plot}
\end{figure*}

\subsection{Baselines and Model Variants}\label{sec:baselines}
We compare our main model described in \Section{sec:method} against the following baselines:
\begin{itemize}
    \item \textbf{\emph{Mean-per-site}:} For each water flow intensity time series $\bs{f}^i=\{f_j^i\}_{j=t_1}^{j=t_{N^i}}$, where $i$ indexes the $i$:th spatial location for a water flow measurement site, we return the mean $\hat{f}^i$ and use that as the predicted water flow intensity at time $t$ (for each day $t$) at the $i$:th site. Note that this provides the optimal prediction (in terms of RMSE) in case only spatial information would be used as model input.
    \item \textbf{\emph{Previous flow}:} Provides $f_{t-1}$ as the predicted water flow intensity at time $t$. Note that this baseline leverages information that our model does not have access to; our model only obtains past rainfall and temperature information, not past water flow intensities.
\end{itemize}

We also train and evaluate the following variants of our proposed ML model:
\begin{itemize}
    \item \textbf{\emph{No-elev}:} This model omits the elevation and terrain slope maps from the set of spatial input maps.
    \item \textbf{\emph{Only-elev}:} For the spatial part of the input, this model only uses the elevation and terrain slope maps. It omits the other spatial input layers.
    \item \textbf{\emph{No-soil}:} This model omits the soil information spatial layers (soil type, soil moisture, soil depth, land cover) from the set of spatial input maps.
    \item \textbf{\emph{Half-time-history:}} Uses temporal information from the past $T=10$ days (instead of $T=20$ as is default).
    \item \textbf{\emph{No-temp}:} This model does not use temperature information as a model input.
    \item \textbf{\emph{No-rain}:} This model does not use rainfall information as a model input.
    \item \textbf{\emph{Flow-(t-k)}:} In addition to all the spatial and temporal inputs of our main model, this model has water flow intensity information $f_{t-T-k+1}, f_{t-T-k+2}, \dots, f_{t-k}$ as an additional temporal input when predicting the water flow intensity $f_t$ at time $t$. We train and evaluate models with $k \in \{1,2,3\}$, i.e.~models that have temporal information up to between three and one day prior to the day for which flow intensities are predicted.
\end{itemize}

Finally, we also train and evaluate the effect of variations to the main model architecture (cf.~\Section{sec:method}):
\begin{itemize}
    \item \textbf{\emph{Alt-rain-temp}:} Uses a more efficient temporal input representation, which results in the input $\bs{I}$ having dimension $H \times W \times (C+T)$ instead of $H \times W \times (C+2T)$. This is achieved by having two unique values per temporal layer (instead of only one), where every second element (spatially) is a rainfall measurement, and every second element is a temperature measurement. Note that the convolutional filters (even the first one) will have sufficiently receptive fields to observe all relevant temporal inputs in this case as well.
    \item \textbf{\emph{FC-early}:} Instead of performing a concatenation of the raw temporal data along the channel dimension, this model first processes the temporal inputs through two fully connected (FC) layers, the last of which produces a $20,000$-dimensional vector. It then reshapes this vector into size $H \times W \times C^\text{temp}=100 \times 100 \times 2$ and concatenates with $\bs{I}^\text{spat} \in \mathbb{R}^{H \times W \times C}$. This data volume of dimension $H \times W \times (C+C^\text{temp})$ is then run through all the layers of the FCN, as for the main model.
    \item \textbf{\emph{FC-mid}:} Similar to \emph{FC-early}, this model first processes the temporal inputs through two FC layers, but here the resulting vector has dimension $2888=38 \cdot 38 \cdot 2$. It then reshapes this vector into size $H^\text{mid} \times W^\text{mid} \times C^\text{temp}=38 \times 38 \times 2$. Different to \emph{FC-early}, this model does not perform the concatenation with the raw spatial data $\bs{I}^\text{spat} \in \mathbb{R}^{H \times W \times C}$; instead it first processes $\bs{I}^\text{spat}$ through the first third of the convolutional layers of the FCN. It then performs the concatenation at this stage of the FCN, followed by joint processing for the remaining two thirds of the network.
\end{itemize}

\subsection{Empirical Results}\label{sec:results}
As mentioned in \Section{sec:training}, we randomly set aside 9 of the 12 data locations for training and 3 for validating the models. The results of our experiments on the validation set are shown in \Figure{fig:main-plot} and \Table{table:main-results} - \ref{table:results-arch}. The evaluation metric that we report is the root-mean-square error (RMSE).

\begin{table}[t]
\centering
\caption{Comparison to models which obtain past water flow intensities as input. Column 1 is our main model that does not obtain previous flows as input. Providing past flow information improves prediction accuracy significantly, but note that in many cases such information is not available.}\label{table:results-flow-input}
\scalebox{1.23}{
\begin{tabular}{|c||c|c|c|}
\hline
\textbf{No flow} & \textbf{Flow-(t-3)} & \textbf{Flow-(t-2)} & \textbf{Flow-(t-1)} \\ \hline \hline
1.35 & 0.94 & 0.80 & 0.63 \\ \hline
\end{tabular}}
\end{table}

Due to the small size of the overall dataset (12 distinct locations), we have not yet considered a proper train-val-test split of the data. This will be done once more data of the appropriate type has been acquired. Currently however, when comparing other model variants and baselines to our main model, we report in the tables the \emph{best} validation RMSE that was obtained during training of the respective model variant. Different to many of the alternative approaches, however, our main model's RMSE on the validation set monotonically improves throughout training (see \Figure{fig:main-plot}), and hence the results of the main model have not been 'cherry picked' at a certain optimal iteration number based on the validation set. Thus any reported improvements of the main model relative to alternatives may in fact be larger if assessed on a withheld test set.
\\ \\
\textbf{Main results.} It can be seen in \Table{table:main-results} that our main model outperforms its ablated variants \emph{no-elev}, \emph{only-elev}, \emph{no-soil}, \emph{no-temp} and \emph{no-rain}. In particular, the elevation and terrain slope maps are crucial, as is past rainfall information. Past temperature information is not as important, but omitting it still results in a higher error. Using rain and temperature information from the past $T=10$ (instead of $T=20$; see \emph{half-time-hist}) days leads to similar results for this data. 

Furthermore, our main FCN method is significantly better than the \emph{mean-per-site} baseline, which indicates that our model has learnt to properly leverage spatio-temporal information. Our approach does however not outperform \emph{previous flow}, which is a very strong baseline that leverages past water flow intensity information. Such information is not available to our model, and is often hard to come by in practical scenarios. Model variants which obtain previous water flow intensity information are however evaluated in \Table{table:results-flow-input}.

\begin{figure*}[ht]
    \centering
\includegraphics[width=0.925\textwidth]{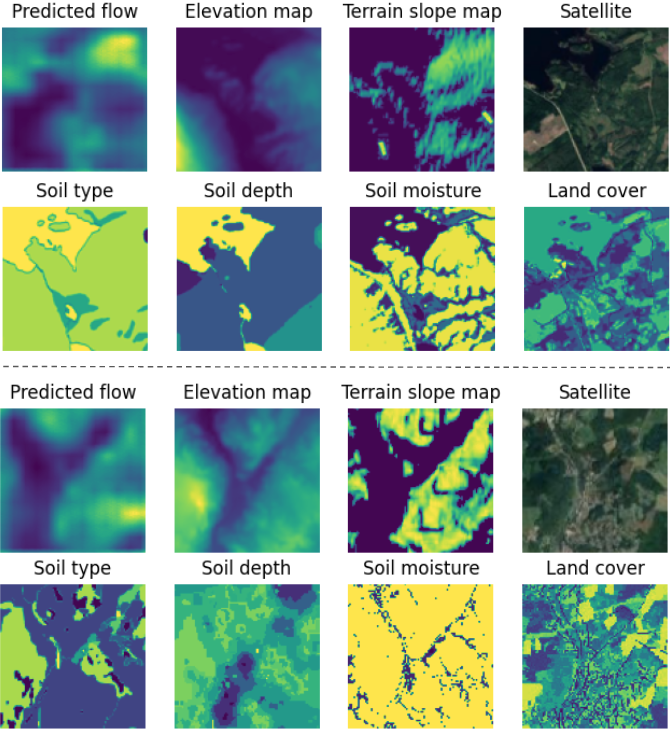}
    \caption{Two qualitative examples for our main model on the validation set (examples differentiated by the dashed horizontal line). In each example, the top-left image represents the predicted water flow intensities at the given area; darker blue means lower intensity, while brighter yellow means higher intensity. The other seven images represent various spatial input layers to the FCN model. For all images except the satellite image, the maximum color intensity is individually normalized so that variations within images become as visible as possible. As can be expected, in both examples, higher flow intensities are typically predicted where the terrain slope is higher. In the example above the dashed line, the model also predicts relatively high flow intensities on the lake that can be observed at the top-left of the satellite image. Note however that the training set contains no ground truth water flow intensities on lakes, and thus the model has never been able to adapt to what is reasonable in terms of flow intensity on lakes.}
    
    \label{fig:viz1}
\end{figure*}

\begin{figure*}[h]
    \centering
\includegraphics[width=0.925\textwidth]{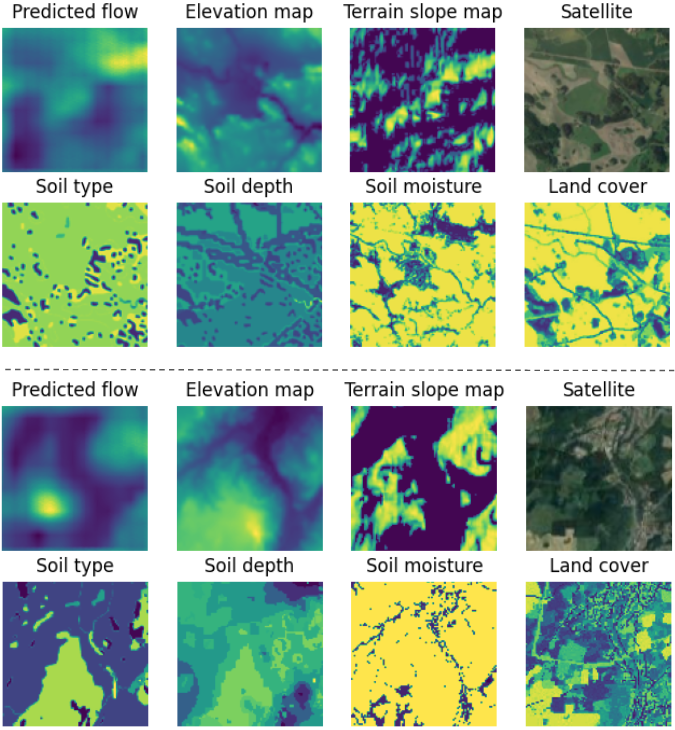}
    \caption{Two additional qualitative examples for our main model on the validation set (examples differentiated by the dashed horizontal line). In the example above the dashed line, the predicted flow is moderately high within most of the map. A peak in terms of predicted flow can be seen at a corresponding peak within the terrain slope map. In the example below the dashed line, in can again be seen that the predicted flow is typically relatively higher where the terrain slope map is higher}
    \label{fig:viz2}
\end{figure*}

In \Figure{fig:main-plot} we show training and validation RMSE curves during model training for our main FCN model. Note that the validation RMSE curve flattens (marginally decreases) throughout training, i.e.~the model does not begin to overfit on the training data despite the relatively small size of the training set.
\\ \\
\textbf{Effect of providing previous water flow intensity information as model input.} As seen in \Table{table:results-flow-input}, models that receive flow information for $T=20$ past consecutive days until three (\emph{flow-(t-3)}), two (\emph{flow-(t-2)}), or one (\emph{flow-(t-1)}) day before the prediction day are significantly more accurate at predicting water flow intensity. Note however that in many practical scenarios such information is not available.
\\ \\
\textbf{Effect of loss function.} In \Table{table:results-loss} we compare the effect of using different loss functions during training; cf.~\eqref{eq:loss}. It is clear that the Huber loss (with $\delta=1.0$ or $\delta=1.1$) yields the best results, whereas the L1-loss results in the worst results.
\\ \\
\textbf{Effect of model architecture.} In \Table{table:results-arch} we compare the effect of using different model architectures. The more efficient \emph{alt-rain-temp} architecture yields almost as good results as our main architecture, so it would be suitable to consider if compute is a limiting factor. The \emph{FC-early} and \emph{FC-mid} architectures yield significantly worse results.
\\ \\
\textbf{Qualitative examples.} Several qualitative examples for our main model are shown in \Figure{fig:viz1} - \ref{fig:viz2}. As can be expected, higher water flow intensities are typically predicted where the terrain slope is high.

\begin{table}[t]
\centering
\caption{Loss analysis. The Huber loss yields the lowest RMSE, with $\delta=1.0$ and $\delta=1.1$ being best. The Huber loss outperforms the MSE loss, and the L1 loss yields poor results.}\label{table:results-loss}
\scalebox{1.15}{
\begin{tabular}{|c|c|c|c|c|}
\hline
\textbf{Huber-1.0} & \textbf{MSE} & \textbf{L1} & \textbf{Huber-0.8} & \textbf{Huber-1.1} \\ \hline \hline
1.35 & 1.55 & 3.17 & 1.47 & 1.35 \\ \hline
\end{tabular}}
\end{table}
\begin{table}
\centering
\caption{Model architecture comparisons. The \emph{FC-EARLY} and \emph{FC-MID} architectures yield significantly worse results than the main model and the \emph{ALT-RAIN-TEMP} architecture.}\label{table:results-arch}
\scalebox{1.15}{
\begin{tabular}{|c|c|c|c|}
\hline
\textbf{Main model} & \textbf{Alt-rain-temp} & \textbf{FC-early} & \textbf{FC-mid} \\ \hline \hline
1.35 & 1.47 & 2.57 & 2.49  \\ \hline
\end{tabular}}
\end{table}


\section{DISCUSSION AND CONCLUSIONS}\label{sec:conclusion}
In this work we have introduced a fully convolutional approach for dense water flow intensity prediction in catchment areas. Our specific results were shown for Swedish basins, but the general methodology is expected to be transferable to other geographical regions.

The proposed model is able to learn and generalize from a limited training dataset. In this work, we have used training data from merely 25 measurement points (28 in total; 3 were used for evaluation). The fact that we obtain such high performance may be attributed to the training setup. In particular, the model generalization is alleviated by the fact that the model sees many slight variations of each measurement site during training, since there are many ways to select a viewpoint around a given measurement site (cf.~\Figure{fig:positions}). To the best of our knowledge, this is the first work which models water flow intensity using a fully convolutional neural network, which allows us to provide \emph{dense} flow predictions -- in effect, we predict one flow intensity per coordinate, even though we only have annotations for 28 specific coordinates. 

Since our main FCN method is significantly better than the \emph{mean-per-site} baseline, we conclude that our model has learnt to properly leverage spatio-temporal information. Our approach does however not outperform the \emph{previous flow} baseline, which is a very strong baseline that leverages past water flow intensity information. Such information is not available to our model, and is often hard to come by in practical scenarios. As can be seen in \Table{table:results-flow-input}, the FCN model variants which utilize past flow information also obtain better results. A potential avenue of future work is thus to consider our setup through a privileged learning lens, wherein flow information could be leveraged during training, but where the model must perform inference using only rainfall and temperature information (in addition to spatial information). 

We hope that our work will serve as a solid stepping stone and an inspiration for further research within dense water flow modeling, which in turn could deliver useful information when it comes to future climate adaptation planning (e.g.~within flood risk management) in Sweden and beyond.

\addtolength{\textheight}{-12cm}   







\bibliography{krycklan-arxiv.bib}
\bibliographystyle{plain}

\end{document}